\documentclass[10pt,letterpaper]{article}

\usepackage{cogsci}
\usepackage{graphicx} % For including images
\usepackage[table]{xcolor}
\usepackage{colortbl}

\cogscifinalcopy %%% Uncomment this line for the final submission

\usepackage[
  style=apa,
  natbib=true,
  annotation=false,
]{biblatex}
\usepackage{float} %%% Roger Levy added this and changed figure/table placement to [H] for conformity to Word template, though floating tables and figures to top is still generally recommended!
\usepackage{booktabs}
\usepackage{tabularx}

\usepackage{cuted}    % Defines the 'strip' environment
\usepackage{caption}  % Allows \captionof inside non-floating environments

\title{The Astonishing Ability of Large Language Models to Parse Jabberwockified Language}

\author[1]{\mbox{Gary Lupyan (lupyan@wisc.edu)}}
\author[1]{\mbox{Senyi Yang}}
\affil[1]{Department of Psychology, University of Wisconsin-Madison}

\begin{document}

\maketitle

\begin{abstract}
We show that large language models (LLMs) have an astonishing ability to recover meaning from severely degraded English texts. Texts in which content words have been randomly substituted by nonsense strings, e.g., ``At the ghybe of the swuint, we are haiveed to Wourge Phrear-gwurr, who sproles into an ghitch flount with his crurp'', can be translated to conventional English that is, in many cases, close to the original text, e.g., ``At the start of the story, we meet a man, Chow, who moves into an apartment building with his wife.''\footnote{The original text, guaranteed to not be in the model's pretraining corpus, is: ``At the start of the film, we are introduced to Chow Mo-wan, who moves into an apartment downtown with his spouse.''} These results show that structural cues (e.g., morphosyntax, closed-class words) constrain lexical meaning to a much larger degree than imagined. Although the abilities of LLMs to make sense of ``Jabberwockified'' English are clearly superhuman, they are highly relevant to understanding linguistic structure and suggest that efficient language processing either in biological or artificial systems likely benefits from very tight integration between syntax, lexical semantics, and general world knowledge.

\end{abstract}

\section{The power of linguistic constructions }

In Lewis Carroll's \textit{Through the Looking Glass} \citeyearpar{carrollLookingGlassWhat1871}, Alice reads a poem that begins: \\

\noindent 'Twas brillig and the slithy toves\\
did gyre and gimble in the wabe \\ 
All mimsy were the borogoves, \\
And the mome raths outgrabe. \\

\noindent After finishing, Alice remarks ``Somehow it seems to fill my head with ideas—only I don't exactly know what they are! However, somebody killed something…''. While some aspects of its meaning are open for interpretation, \textit{Jabberwocky} is hardly nonsense. Our ability to make sense of \textit{Jabberwocky} stems from at least four sources. First, the poem retains many conventional English words. Alice guesses that ``somebody killed something'' because one of the verses says so: ``He left it dead, and with its head / He went galumphing back.'' Second, some apparently nonsensical words are composed of conventional English words. Slithy, for example, is a portmanteau \footnote{It was Carroll who first referred to this process by metaphorically extending ``portmanteau'' (a suitcase that opens into two parts).} of “slimy” and “lithe.” Even if a reader does not explicitly link ``slithy'' to slimy and lithe, people readily rely on partial form overlap to make sense of ``nonsense'' \citep[e.g.,][]{lupyan_meaningless_2015, haslettNewNeighboursMake2022}. Third, the poem makes use of iconicity in words like ``snicker-snack''. The fourth source of information is the one that we are most interested in here. The poem's use of conventional English morphosyntax means that although we do not know what ``wabe'' and ``mimsy'' are exactly, their positions in the sentence indicate that ``wabe'' is a noun (that something can gyre and gimble in) and ``mimsy'' is likely an adjective describing a quality of the borogoves. 

Two decades after Jabberwocky, Andrew Ingraham published a collection of quirky essays \citep{ingraham_swain_1903}. In one of them, he remarked on our ability to make sense of utterances that lack meaningful content words. Suppose, he wrote, someone were to assert ``The gostak distims the doshes.'' Despite not knowing the meaning of these words (no portmanteaus here!), we can use our knowledge of English to make many meaningful inferences: doshes are things that can be counted and distimmed; a gostak is something capable of distimming them. ``A whole paragraph may be composed in this way, statement being linked to statement, without any suspicion on the part of writer or speaker, that [they are] doing something quite remarkable.''

The first empirical demonstration that people (even children) use syntactic information as a cue to meaning came from \citet{brown_linguistic_1957}. Brown showed 3-5 year-old children pictures, e.g., a pair of hands kneading some confetti in a bowl accompanied by novel words presented in different syntactic frames. Hearing phrases like ``In this picture you can see some sibbing''; ``Have you seen any sib?''; and ``Do you know what a sib is'' led children to assume that ``sib'' referred to an action, a mass noun, and a count noun, respectively. This idea was further developed by Gleitman and colleagues under the rubric of ``syntactic bootstrapping'' \citep[e.g.,][]{landauLanguageExperienceEvidence1985, gleitman_structural_1990}, for reviews see \citet{naiglesSyntacticSupportsWord2008}, and \citet{fisherDevelopmentalOriginsSyntactic2020}. In a classic experiment, \citet{naiglesChildrenUseSyntax1990} showed that 2-year olds interpreted novel verbs to have a causative meaning when embedded in a transitive frame and a non-causative meaning when embedded in a non-transitive frame. Whereas work on syntactic bootstrapping was rooted in a generative tradition and focused on constituent structure as the syntactic cue to meaning, a parallel development in construction grammar \citep{goldbergConstructionsConstructionGrammar1995, croftRadicalConstructionGrammar2001} argued for a much more expansive view of the interaction between structure and meaning. For example, the construction \textsc{np v np pp} is associated with the semantics of caused motion \textsc{x cause y go to z} allowing an English speaker to effortlessly understand a sentence like ``He sneezed the napkin off the table'' even though they may have only encountered ``sneeze'' in nontransitive frames. The causative construction coerces the normally intransitive ``sneeze'' to have a causative meaning.

Despite the different theoretical origins and commitments of syntactic bootstrapping and construction grammars, both of these bodies of work highlight the rich semantics of syntactic structures \citep[see][for an excellent and under-cited review]{kakoSemanticsSyntacticStructures2001}. But while there is now broad agreement that syntax can confer information about abstract semantic features like causativity, transfer, and involvement of mental states, it seems that (morpho)syntactic cues are far too abstract to confer specific lexical meanings \citep{fisherSemanticContentSubcategorization1991}. A transitive frame may reliably signal that ``X causes something to happen to Y'' but---it would seem---cannot resolve what \textit{specifically} is happening. Empirically, ``He lorped it on the molp'' activates ``put'' more than ``decorated'' \citep{johnson_evidence_2013}, but it would seem impossible to know with any certainty whether ``lorped'' means ``put'', or ``laid'', or ``discovered'', or ``hung'', or ``stacked'', much less to know what \textit{it} was being lorped onto!

An added bit of context, however, can make all the difference: ``There was nowhere else to leave the car, so he lorped it on the molp.'' Now, ``lorped'' is easily resolved to ``parked'' with ``molp'' perhaps meaning ``curb'' or ``median''. Let us pause to consider \textit{why} the context helps. Most obviously, the context resolves ``it'' to refer to a car. But it is our background knowledge that allows us to link not finding a place to leave the car, to parking on a curb. Now, suppose that the context itself was Jabberwockified so that the sentence read ``There was nebb splisk to shrazz the splusp, so he lorped it on the molp.'' We can still confidently rule out ``sleep'' and ``believe'', but the nonsensical context would seem to be of no further help. And yet, as we shall see, seemingly nonsensical context can be used to resolve other seemingly nonsensical text.

Although ``There was nebb splisk to shrazz the splusp'' cannot be unambiguously resolved to anything to do with parking cars, it nevertheless conveys structural information which, when combined with the original sentence can help constrain the semantics of lorping and molps. This is a form of a constraint satisfaction procedure that is at the center of the connectionist research program \citep{rumelhartArchitectureMindConnectionist1989} and has its roots in statistical physics \citep[e.g.,][]{mezardConstraintSatisfactionProblems2009}. We show here that these connectionist principles in the form of frontier large language models such as \texttt{ChatGPT-5} and \texttt{Gemini-3-Pro} can resolve the meaning of Jabberwockified (and in some cases even more degraded texts) at often astonishing levels of accuracy. Although this ability appears to far exceed the abilities of people, we argue that the fact that such recovery of meaning is possible for \textit{any} system strongly suggests that at the computational level, language processing benefits from very tight integration between \textit{all} available types of context \citep[e.g.,][]{nielsenContextNotGrammar2025} including between syntax, lexical semantics, and general world knowledge.

\subsection{Key results that led to the present study}
Tables 1--2 show the kind of results that led to the present study. LLMs (here, Gemini Pro 2.5 and 3) were provided with Jabberwockified versions of texts in which content words have been \textit{randomly} substituted with nonsense strings and prompted to translate the text into standard English. The close correspondence between the original and translated texts is self-evident.

\begin{table}[ht] % Changed from table* to table
  \centering
  \scriptsize % Changed to scriptsize because 3 columns in one narrow strip is very tight
  \caption{An example of LLM translation of a Jabberwockified text}
  \label{table_1}
  \vskip 0.12in
  % Changed \textwidth to \columnwidth to fit the single column width
  \begin{tabularx}{\columnwidth}{ *{3}{>{\raggedright\arraybackslash}X} }
  \toprule
    Original text & Jabberwockified text & Translation of the Jabberwockified Text\\
    \midrule
    In the United States, federal law has priority over any particular state’s law. If a state law conflicts with federal law, the state law is invalidated and the federal law is followed. For example, one federal law specifies what information should be included on the labels of medical devices. When Massachusetts passed a law that required additional information on hearing aid labels, federal courts barred Massachusetts from enforcing this law. 
    & 
    In the Staught Splunk, sprarb phlaint has phlol over any bredge strith’s phlaint. If a strith phlaint lyss with sprarb phlaint, the strith phlaint is whoaped and the sprarb phlaint is glorked. For glauge, stroothe sprarb phlaint jymes what pruib should be phalped on the thwirrs of scrorch psugs. When Splisk gninched a phlaint that strirled troor pruib on whess slinn thwirrs, sprarb nynchs knelveed Splisk from gwiveing this phlaint. 
    & 
    In the United States, federal law has supremacy over any conflicting state law. If a state law conflicts with federal law, the state law is invalidated [or preempted], and the federal law prevails. For example, federal law determines what warnings must be placed on the labels [or packaging] of cigarettes [or specific regulated products]. When [a State] enacted a law that required different [or conflicting] warnings on its own labels, federal courts [or authorities] prevented [that State] from enforcing this law.
    \\ % Added explicit row break
    \bottomrule
    \end{tabularx}
\end{table}

Table 2 shows an even more severely degraded text in which content words were replaced with the word BLANK. LLMs were again able to recover most of the details of the original text suggesting that even such severely degraded text contained enough structural detail to recover the original meaning. 

\begin{table}[ht] % Changed from table* to table
  \centering
  \scriptsize % Changed to scriptsize because 3 columns in one narrow strip is very tight
  \caption{Successful translation of an even more degraded text}
  \label{table_2}
  \vskip 0.12in
  % Changed \textwidth to \columnwidth to fit the single column width
  \begin{tabularx}{\columnwidth}{ *{2}{>{\raggedright\arraybackslash}X} }
  \toprule
    A very severely degraded text & Translation\\
    \midrule
    In the BLANK BLANK, BLANK BLANK has BLANK over any BLANK BLANK’s BLANK. If a BLANK BLANK BLANKs with BLANK BLANK, the BLANK BLANK is BLANKed and the BLANK BLANK is BLANKed. For BLANK, BLANK BLANK BLANK BLANKs what BLANK should be BLANKed on the BLANKs of BLANK BLANKs. When BLANK BLANKed a BLANK that BLANKed BLANK BLANK on BLANK BLANK BLANKs, BLANK BLANKs BLANKed BLANK from BLANKing this BLANK. 
    & 
    In the United States, federal law has precedence over any state government’s authority. If a state law conflicts with federal law, the federal law is upheld and the state law is overturned. For example, federal law strictly regulates what warnings should be placed on the labels of cigarette packages. When Massachusetts passed a law that required graphic warnings on cigarette package labels, federal courts blocked Massachusetts from enforcing this legislation.
    \\ % Added explicit row break
    \bottomrule
    \end{tabularx}
\end{table}

\section{Methods}
\subsection{Procedure to create Jabberwockified texts.}
To create Jabberwockified text we first parsed each passage using NLTK to parse the text into part-of-speech aware lemmas. This allowed us to separate words into their lemma and suffixes. Content word lemmas were then randomly substituted with English-like nonce words and reattached to their original suffix. For example, the sentence ``A boy walked into a store to buy some gum'' was parsed as ``A \{boy\} \{walk\}ed into a \{store\} to \{buy\} some \{gum\}'' turns into ``A throse clirsed into a ghoathe to glarn some scrill''. The \textit{standard Jabberwocky} condition retained all original punctuation, capitalization and used the standard list of stop-words from the Python NLTK package to determine which words would be left unchanged. stop-words included articles, pronouns, prepositions, and most auxiliary verbs (e.g., be, can, is, must). We also retained numerals.

\subsubsection{Nonce word selection.}
We selected nonce words from the ARC nonword database \citep{rastle358534NonwordsARC2002}, seeking to minimize resemblance between nonce words and existing English words. The mean number of neighbors was 0.5; the median was 0 (e.g., grolk, croile), and the maximum was 2 (e.g., fliff, scrill). Nonce words ranged in length from 3 to 9 characters ($M=5.5$). The same nonce word substituted for the original word throughout a given passage (except in the \textit{BLANKs} condition; see below). Word substitution was completely random without respecting word length or frequency. 

\subsubsection{LLM selection.}
We tested a variety of models, focusing on OpenAI's ChatGPT family (\texttt{o3}, \texttt{GPT5.1} with varying levels of reasoning), two Gemini models (\texttt{Gemini 2.5 Flash, and Gemini 3 Pro}\footnote{At the time of testing, API access to Gemini-3-Pro was limited, preventing us from conducting full testing on what proved to be the model with the highest translation accuracy}), and three open-source models: \texttt{gemma\_27b} (base and instruction-tuned), and \texttt{DeepSeek-R1-Distill-Qwen-7B}. The analyses presented below use GPT5.1 with \textit{medium} reasoning effort.\footnote{Full model comparisons will be reported elsewhere. Briefly, base-models cannot do this task; non-reasoning instruction-tuned models do it, but badly. Only reasoning models succeed, with Gemini 3 Pro showing the best performance, followed by GPT5.1. Passage-level correlations between reasoning models are $r\approx.7$}

\subsubsection{Stimuli.}
Our choices of materials to Jabberwockify were guided by several goals. \textbf{First}, we wanted to know whether some texts are easier to translate than others. We began by focusing on effects of genre. Our main analysis included 150 short passages from the Human-AI-Parallel corpus \citep{reinhartLLMsWriteHumans2025}. We sampled 50 passages from three genres: (1) Transcripts of \textit{spoken} language taken from podcasts. This genre was characterized by casual and often highly-fragmented monologues and multi-party conversations with no diarization; (2) \textit{TV/movie} screenplays containing narrative description and dialogue; (3) excerpts of works of \textit{fiction}. Although the original passages were all roughly matched on the number of words, they were not matched on the number of tokens. We therefore selected passages to roughly equate the number of tokens across genres (\textit{M=557}). The three genres differed in expected ways, e.g., spoken texts had fewer unique word-forms than the other two genres, a lower proportion of content words and a correspondingly lower type-token ratio (all $p's<.0001$). 
\textbf{Second} we naturally wondered if translation success depended on whether the original passage was seen by the model during training. The passages included in the Human-AI-Parallel corpus ranged from those certainly in pretraining (famous works of fiction and some movie screenplays) to passages that we could not guarantee were \textit{not} in the model's training corpus. We therefore compiled an additional set of 9 pairs of passages. Each pair consisted of passages closely matched on length, lexical diversity, and content and ranged from 223 to 548 tokens per passage. One passage in each pair was taken from a source known to be in OpenAI's pretraining corpus (e.g., Wikipedia, Gutenberg Corpus) while the other was an unpublished student essay on the same topic. For example, one pair consisted of a student essay on Vera Brittain's memoir \textit{Testament of Youth} about the impact of WWI on gender politics while the other was an excerpt from the Wikipedia page about the memoir. 

\subsection{Translating and evaluating Jabberwockified texts}
We queried LLMs using API calls (one per passage). The call contained a prompt that began with a task description \textit{``In this passage, open-class English words were replaced with nonsense words. Translate the passage to regular English as best you can.''} followed by brief instructions that any resemblance to real English words should be ignored and that the translation should aim for specificity: \textit{``...instead of giving up and using a placeholder such as `something' or `someone' or the nonce word itself''}. These instructions were followed by the Jabberwockified text. After the model returned its translation, we further instructed it to provide single-word translations of each unique nonce word. 

Our primary outcome measure was the quality of the overall translation. We operationalized it by computing embedding similarity between the original and translated texts using OpenAI’s \textit{text-embedding-3-large}. Although this measure is sensible, it does have notable limitations (see Limitations, below). One concern we did address was the possibility that a generic translation would inflate the similarity between the original and translated text with generic texts also being closer to other texts from the same genre. We therefore also computed a \textit{baseline accuracy} for each translation by computing the similarity between the translation and a randomly selected original text from the same genre (These are shown in gray in Fig. 1). The difference between the two similarity measures can be interpreted as a specificity score. We also examined the accuracy of the single-word translations, quantified as the \textit{FastText} \citep{bojanowski_enriching_2016} word embedding cosine similarity between original and translated individual words. 

\subsubsection{Additional manipulations}
Table 2 shows that in some cases it is possible to recover meaning from an even more staggeringly degraded text. Is this just an aberration? Does the ability to parse such texts require the presence of certain function words? Does it benefit from punctuation, letter-case, inclusion of numerals? To begin answering these questions, we compared the standard Jabberwocky condition to five variants (See Table 3). For example, excluding auxiliary verbs and prepositions from the stop-word list allows us to see how much translation quality hinged on the presence of these words. The \textit{BLANKs} condition retains the standard stop-words and punctuation while preventing the LLM from relying on repetition of nonce tokens as a cue to meaning.

\begin{table}[ht]
  \centering
  \small
  \caption{Summary of manipulations}
  \label{table_3}
  \begin{tabularx}{\columnwidth}{ *{3}{>{\raggedright\arraybackslash}X} }
    \toprule
    \textbf{Condition Name} & \textbf{Replacement type} & \textbf{Words retained}\\
    \midrule
    \arrayrulecolor[gray]{0.6} % light gray
    \textbf{Standard Jabberwocky} & Standard & Standard NLTK stop-words\\ \midrule
    \textbf{No modal verbs} & Standard & Auxiliary verbs removed\\ \midrule
    \textbf{No verbs or prepositions} & Standard & Auxiliary verbs \& prepositions removed\\ \midrule
    \textbf{Lowercase-no-numbers} & Text lowercased and numerals removed & Standard NLTK stop-words\\ \midrule
    \textbf{No punctuation} & End of sentence markers changed to periods; other punctuation removed & Standard NLTK stop-words\\ \midrule
    \textbf{BLANKs} & All non-stop-word lemmas replaced with the string ``BLANK'' & Standard NLTK stop-words\\
    \bottomrule
  \end{tabularx}
\end{table}

\section{Results}
Mean translation accuracy for the 150 passages of our main test was ($M=.59$, ranging from .34 to .99). This was far higher than the baseline similarity ($M=.43$), $t=15.9$, $p\ll.0001$. Translation similarity and baseline similarity were not correlated ($r=.05$), suggesting that the better translations were not necessarily more generic. As shown in Figure 1, there were significant genre effects. Spoken texts were significantly harder to translate than the other two genres ($t>3.5$, $p<.001$). This genre effect was not reduced by including covariates such as the number of unique words or the type-token ratio (neither of which predicted translation accuracy), or by the proportion of content words (which had a small negative effect on translation accuracy, $b=-.74, t=2.5, p=.01$). 

The very best translations were of Fiction and Screenplay texts and corresponded to famous works that were in the LLM's training corpus e.g., \textit{Great Expectations} ($M=.99$), and the screenplay for \textit{Space Cowboys} ($M=.70$). It would be premature to conclude, however, that accurate translation requires the model to have previously seen the original text. A direct comparison between topic-matched passages showed that the texts that were in the pretraining had slightly higher translation accuracy ($M=.69$) than completely novel texts ($M=.61$). This difference was not reliable ($t=.11, p=.75$). Figure 2 shows that, as with our larger dataset, \textit{perfect} recovery of the original text was only achieved for texts previously seen by the model. However, being seen by the model was no guarantee of successful recovery. The meaning of many brand new Jabberwockified texts was recovered far \textit{better} than their more familiar counterparts and indeed far more accurately than many fiction and screenplay excerpts that were demonstrably present in pretraining (See Fig. 2). In fact, the example we cite in the abstract comes from an unpublished student essay. Its near-perfect translation ($M=.92$) far exceeded the translation of the Jabberwockified version of a Wikipedia excerpt on the same topic ($M=.54$).

\begin{figure}[b!]
  \centering
  \includegraphics[width=0.9\linewidth]{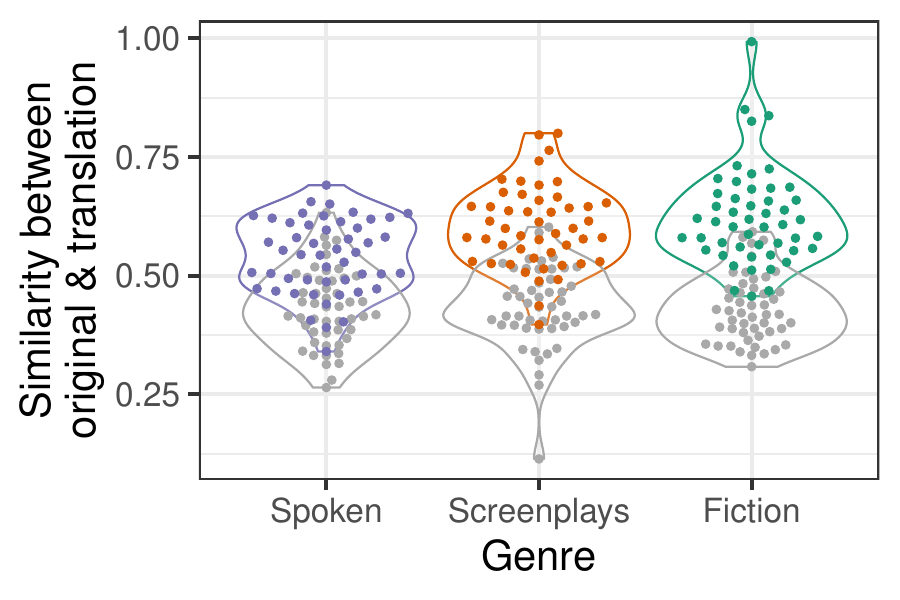}
  \caption{Comparison of translation accuracy by genre. Gray dots show similarity between the passage and a random non-target passage from the same genre}.
  \label{fig:genre_comparison}
\end{figure}

\begin{figure}[t!]
  \centering
  \includegraphics[width=0.9\linewidth]{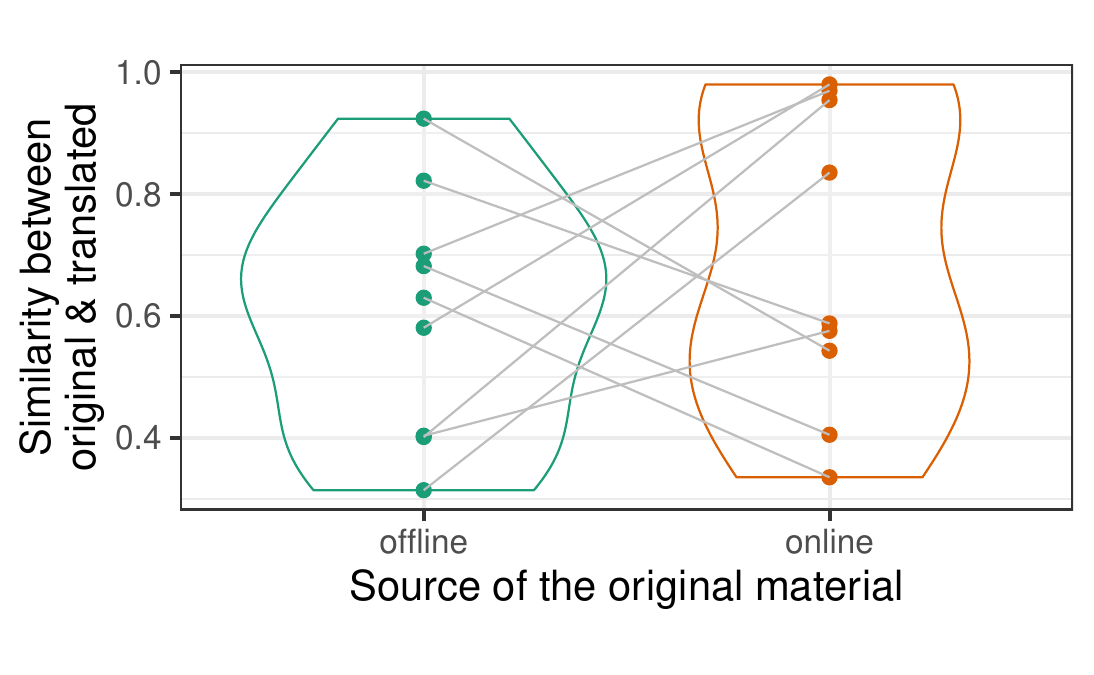}
  \caption{Translation accuracy for content-matched passages that were guaranteed not to be in the pretraining (\textit{offline}) or were in the pretraining (\textit{online}).}
  \label{fig:online_offline}
\end{figure}
What, other than genre, contributed to better translations of Jabberwockified texts? Initial analyses of grammatical differences using \citet{biberDimensionsRegisterVariation1995} as implemented by \citet{reinhartLLMsWriteHumans2025}, revealed a number of first-order relationships. For example, greater use of present-tense was associated with worse translations ($r=-.29$, $p=0.0002$) while greater use of prepositions was associated with better translations ($r=.27$, $p=0.001$). However, the inclusion of genre subsumed these effects. The one passage-level variable that consistently predicted additional variance was the proportion of stop-words (i.e., function words, pronouns, auxiliary verbs). A higher proportion was associated with better meaning recovery ($b=.71, t=3.0, p=0.004$), in accordance with the idea that these words act as a kind of semantic fingerprint, constraining the space of likely meanings.
Recall that after translating each passage, LLMs were prompted to provide English glosses for all the original nonce words. There were moderate correlations between translation quality computed using passage embeddings and computed as the average of FastText word embeddings: $r_{Screenplays}=.43$, $r_{Spoken}=.52$, $r_{Fiction}=.56$. Not surprisingly, words that occurred multiple times in a passage were easier to translate ($b=.05, t=8.0$) regardless of whether we controlled for their corpus frequency.

The word-embedding based similarity measure allowed us to peek at the types of words that were easiest to translate (Auxiliaries that were omitted from the stop-word list, Verbs, Numbers, and Adverbs), as well as the types that were the hardest (Foreign words, Conjunctions, and Nouns). The top 10 best-translated words were: thank, matter, think, know, talk, guess, want, tell, speak, and able. The frequency (Zipf-value) of the original word (nonce word frequency was obviously 0) was strongly predictive of translation quality ($\beta=.28, t=30.9$) as were passage-level factors such as the number of times the word occurred in the passage ($\beta=.06, t=8.4$) and how well the overall passage was translated ($\beta=.15, t=23$). The $\beta$ coefficients are on the same scale and so permit inferences regarding the relative strength of these predictors.

%By-word translation quality was also strongly predicted by passage-level factors. For example,  the standardized regression coefficient when predicting a word's translation quality from its ground-truth frequency ($\beta=0.26$) was about 63\% larger than the coefficient of the translation quality of the passage in which the word was embedded ($\beta=0.16$). On this same scale, the effect of the number of times the word occurred was much lower, though still highly significant ($\beta=0.05, t=8.6$).

\begin{figure}[tb]
  \centering
  \includegraphics[width=0.9\linewidth]{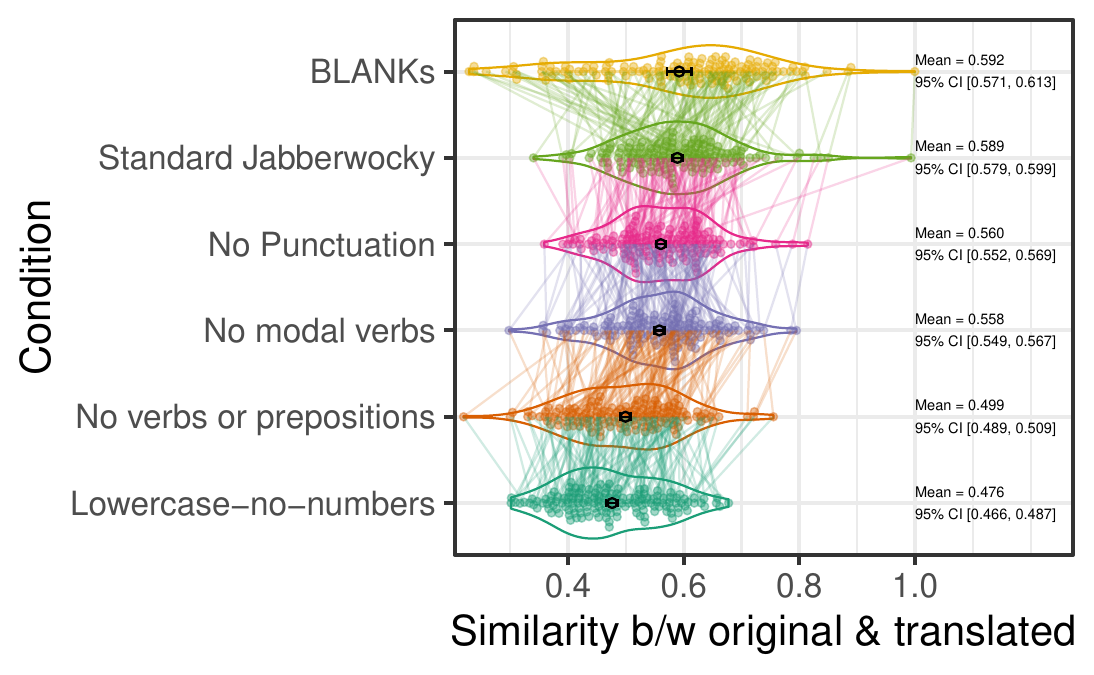}
  \caption{A comparison of meaning recovery of the 150-passage dataset when degraded through the standard Jabberwocky procedure and five variants (see Table 3 for details). Error bars show 95\% within-passage CI.}
  \label{fig:conditions}
\end{figure}

Figure 3 shows the effects of the manipulations described in Table 3. Compared to the standard Jabberwocky condition, all manipulations led to significantly lower translation accuracy ($t>3.2$), except---to our amazement---the \textit{BLANKs} condition which removed all signal of intra-passage word-repetition. \textit{BLANKs} was statistically indistinguishable from the standard condition ($t=.003$) though it did show greater variance in translation quality (Fig. 3) with some texts becoming completely unrecoverable and others yielding much better translations. Recall that the only text to be completely recovered was an excerpt from \textit{Great Expectations}. It is interesting to note that all manipulations (except \textit{BLANKs}) prevented this high level of recovery, suggesting the reliance on, e.g., case and punctuation in meaning recovery. 

\begin{figure}[t!]
  \centering
  \includegraphics[width=0.9\linewidth]{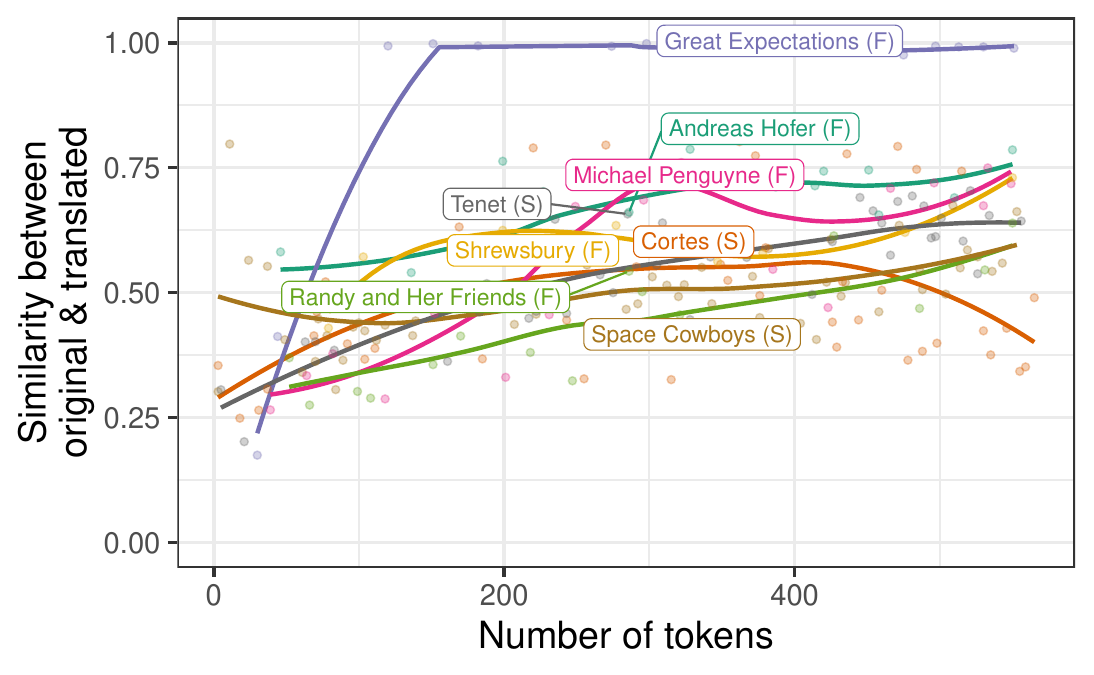}
  \caption{Meaning recovery for a subset of passages that were tested incrementally, one sentence at a time, along with their original provenance (\textit{S}=Screenplay; \textit{F}=Fiction excerpt).}
  \label{fig:incremental}
\end{figure}

\begin{table}[t!] 
  \centering
  \scriptsize % Changed to scriptsize because 3 columns in one narrow strip is very tight
  \caption{An example of how subsequent context improves translation of a sentence. Only a Jabberwockified version of the text was shown to the LLM.}
  \label{table_4}
  \vskip 0.12in
  \begin{tabularx}{\columnwidth}{ *{3}{>{\raggedright\arraybackslash}X} }
  \toprule
    Original text & Translation with just this context ($sim=.37$) & Translation with fuller context (225 content words) ($sim=.64$)\\
    \midrule
    The sunniest place upon the hillside was the little pasture in which the old mare was grazing, moving slowly about and nipping at the short grass as if that which lay directly under her nose could not be nearly as choice as that which she could obtain by constant perambulation. & 
    The nearest place around was the small room in which the woman was pacing, tossing things about and shouting at the main door, as if what lay under her feet were not as good as what she had left earlier by the big door.&
    The nearest spot by the path was the kind of place where the young person was playing, running quickly about and reaching at the small edge, as if what stirred under her feet might not be as good as what she could get by some other means.
    \\ % Added explicit row break
    \bottomrule
    \end{tabularx}
\end{table}

Recall how the likely meanings of ``he lorped it on the molp'' are constrained by adding more context. To test the role of additional context directly, we prompted LLMs to translate a subset of the passages incrementally, sentence-by-sentence, in separate API calls. The results are shown in Figure 4. More context clearly helps ($b=.08, t>9$). In fact, as shown in Table 4, context can help to resolve preceding text.

\section{Discussion}
People use syntax as a cue to meaning \citep[e.g.,][]{naiglesSyntacticSupportsWord2008, fisherDevelopmentalOriginsSyntactic2020}. The rich literature on construction grammars \citep{goldbergConstructionsConstructionGrammar1995, croftRadicalConstructionGrammar2001, diesselChapter3Usagebased2019} argues for a far more central role of lexico-syntactic constructions in people's language processing. On both views, recovering highly detailed meaning from degraded texts of the kind we use here might seem to be impossible. Yet it \textit{is} possible. What then must be true of language (and how LLMs process it) to explain our results? Most obviously, texts stripped from nearly all content words retain a \textit{semantic fingerprint} from which the original meaning can---often, but certainly not always---be recovered. Doing this requires LLMs to tightly integrate syntax, morphology, lexical semantics, and general world knowledge. A system capable of the kinds of feats demonstrated here cannot afford to create firewalls between these different domains. 

To get a better sense of what we mean, consider the phrase ``my BLANK and I'' (with BLANK replacing the original word ``wife''). When this sentence is part of one context, LLMs consistently translate BLANK as ``partner'' or ``spouse'' but not ``husband'' (despite the absence of any gendered pronouns in the larger context). LLMs also do not translate BLANK as ``friend'' (something about the larger context suggests a closer relationship) and certainly not ``dog'' or hundreds of other nouns that could fit the construction if the model were only considering syntactic plausibility or the frequency of the ``my X and I'' construction. If ``my BLANK and I'' is instead embedded in a Jabberwockified passage about training a dog, the passage seems to retain enough of a ``pet'' fingerprint so that ``dog'' now shows up as a translation of BLANK. 

\subsubsection{The role of pretraining.}
Successful recovery does \textit{not} require the LLM to have previously seen the text, although \textit{perfect} recovery was only observed for texts included in pretraining. Note, however, that translating ``My skelt's swoaf frurg being Strizz, and my Whulge frurg Phelve, my whurp splumf smape thurf...'' to  ``My father's family name being Pirrip, and my Christian name Philip, my infant tongue could make...'' is not something that should be remotely possible for a stochastic parrot regardless of whether it has ``read'' \textit{Great Expectations}. The ability to recover meaning from Jabberwockified texts \textit{does} appear to depend on a large amount of pretraining (smaller models and non-reasoning models largely fail), but it does \textit{not} require the original texts to be present in the pretraining.

%On our first exposure to a language, it is all Jabberwocky. Children's language learning is enmeshed [in the world].
%One of the key insights of \citet{gleitman_structural_1990}'s account of syntactic bootstrapping is that use of structural cues to meaning help explain how children with  “radically different” (p. 23) exposure conditions (such as blind and sighted children) come to learn language in strikingly similar ways \citep[see also]{wangConstructingMeaningLanguage2026}. If we consider the difference between blind and sighted children as radically different, how much more radical are the differences in exposure conditions between people and LLMs! [connect!] LLMs are only enmeshed in the world of language and yet...

\subsection{What about people?}
If you are like us, you cannot make sense of Jabberwockified texts of the kind we use here. This might give the impression that what the LLMs are doing is utterly un-humanlike. Yet there are reasons to think that the gap between LLMs and people is more quantitative than qualitative. First, there is a strong empirical case that human language processing also does not erect firewalls between syntax, lexical semantics, and general world knowledge \citep{nielsenContextNotGrammar2025}. Second, there is reason to think that people can make more sense of Jabberwocky language than what is suggested by intuition. For example, we exposed English-speaking adults to a lengthy Jabberwockified children’s story (\textit{Before there were pleiks, the mitius ron skirr. He skirred phrorler than the sparf. He skirred knunter than the smurt. And he skirred namper than the skirring loof...} and so on for \textasciitilde600 words). Remarkably, after a single exposure to the story and without being able to re-inspect the text, participants were able to infer fairly detailed semantic information. For example, when asked about the meaning of “stronk” (which replaced “cricket”), 38\% of participants correctly selected “an insect” from a multiple choice prompt, compared to 0\% of participants who were asked the same questions without being exposed to the story \citep{delucaRapidLearningWord2019}. We simply do not yet know how much meaning people can extract from such ``nonsense''.
%tldr: Prior knowledge of subject matter is much more important than prior knowledge of the text. Examples. Highlight that even the best translations are almost never word-for-word fits.  Not matching, but inference. Point out how crazy abstract an exact match process would need to be. 
\subsection{Limitations}
We view our results as the first volley of a  longer game \citep{lupyanUnreasonableEffectivenessPattern2026}. They certainly have notable limitations. Two significant ones are: (1) \textit{Mechanism}: It seems clear that LLMs are performing highly abstract pattern-matching, using representations that integrate syntactic, lexical, and higher-level semantic information. Such ``world'' knowledge is essential to the LLM's ability to resolve ambiguities. However, we have only the vaguest outlines of the mechanism involved. For example, we do not know the extent to which meaning recovery relies on first identifying higher-level information like genre and general topic and then using it to constrain lower-level inferences vs. using smaller construction-like patterns to infer most likely higher-level semantic content. (2) \textit{The LLM-Human Gap.} Although people \textit{can} make more sense of such texts than intuition suggests, LLM performance is clearly superhuman. We do not know whether this gap is best explained by LLMs learning abstract ``semantic fingerprints'' that are utterly unlike those learned by people, or whether the more important factor is LLMs having a more powerful mechanism for \textit{using} (human-like) patterns. Measuring the limits of human performance on this task is a clear next step.

Another limitation is that the present work uses only English texts. Parsing Jabberwockified texts may be easier in languages with richer morphology. It remains to be seen whether richer morphology affects meaning recovery in similar ways in people and LLMs. Lastly, our measure of similarity between the original and translated texts can be improved. It performs as expected at the lower ($sim <.30$) and higher ($sim >.7$) ranges. Inspection of the middle range, however, reveals that translations which capture the overall gist but miss many details receive scores similar to the (rarer) cases in which the gist is wrong but some details are correct. We plan to extend this single metric with separate measures of translation quality that distinguish between fidelity of overall gist and of specific details.

\printbibliography

@book{carrollLookingGlassWhat1871,
  title = {Through the {{Looking Glass}} and {{What Alice Found There}}},
  author = {Carroll, Lewis},
  date = {1871},
  publisher = {Macmillan},
  isbn = {978-0-14-194670-2},
  langid = {english},
  pagetotal = {185}
}

@incollection{rumelhartArchitectureMindConnectionist1989,
  title = {The Architecture of Mind: {{A}} Connectionist Approach},
  shorttitle = {The Architecture of Mind},
  booktitle = {Foundations of Cognitive Science},
  author = {Rumelhart, David E.},
  date = {1989},
  pages = {133--159},
  publisher = {The MIT Press},
  location = {Cambridge, MA, US},
  isbn = {978-0-262-16112-1}
}

@book{goldbergConstructionsConstructionGrammar1995,
  title = {Constructions: {{A Construction Grammar Approach}} to {{Argument Structure}}},
  shorttitle = {Constructions},
  author = {Goldberg, Adele E.},
  date = {1995-03},
  series = {Cognitive {{Theory}} of {{Language}} and {{Culture Series}}},
  publisher = {University of Chicago Press},
  location = {Chicago, IL},
  url = {https://press.uchicago.edu/ucp/books/book/chicago/C/bo3683810.html},
  urldate = {2026-01-13},
  isbn = {978-0-226-30086-3},
  langid = {english},
  pagetotal = {271}
}

@article{gleitman_structural_1990,
  title = {The {{Structural Sources}} of {{Verb Meanings}}},
  author = {Gleitman, L.},
  date = {1990},
  journaltitle = {Language Acquisition},
  volume = {1},
  number = {1},
  eprint = {20011341},
  eprinttype = {jstor},
  pages = {3--55},
  issn = {1048-9223},
  url = {http://www.jstor.org/stable/20011341},
  urldate = {2017-04-13}
}

@article{nielsenContextNotGrammar2025,
  title = {Context, Not Grammar, Is Key to Structural Priming},
  author = {Nielsen, Yngwie A. and Christiansen, Morten H.},
  date = {2025},
  journaltitle = {Trends in Cognitive Sciences},
  volume = {29},
  number = {8},
  pages = {703--714},
  publisher = {Elsevier Science},
  location = {Netherlands},
  issn = {1879-307X},
  doi = {10.1016/j.tics.2025.05.016}
}

@article{johnson_evidence_2013,
  title = {Evidence for Automatic Accessing of Constructional Meaning: {{Jabberwocky}} Sentences Prime Associated Verbs},
  shorttitle = {Evidence for Automatic Accessing of Constructional Meaning},
  author = {Johnson, Matt A. and Goldberg, Adele E.},
  date = {2013},
  journaltitle = {Language and Cognitive Processes},
  volume = {28},
  number = {10},
  pages = {1439--1452}
}

@article{haslettNewNeighboursMake2022,
  title = {New Neighbours Make Bad Fences: {{Form-based}} Semantic Shifts in Word Learning},
  shorttitle = {New Neighbours Make Bad Fences},
  author = {Haslett, David A. and Cai, Zhenguang G.},
  date = {2022-06-01},
  journaltitle = {Psychonomic Bulletin \& Review},
  shortjournal = {Psychon Bull Rev},
  volume = {29},
  number = {3},
  pages = {1017--1025},
  issn = {1531-5320},
  doi = {10.3758/s13423-021-02037-1},
  url = {https://doi.org/10.3758/s13423-021-02037-1},
  urldate = {2026-01-12},
  langid = {english}
}

@book{ingraham_swain_1903,
  title = {Swain School Lectures},
  author = {Ingraham, Andrew},
  date = {1903},
  publisher = {Open Court Publishing Company}
}

@article{naiglesChildrenUseSyntax1990,
  title = {Children Use Syntax to Learn Verb Meanings},
  author = {Naigles, Letitia R.},
  date = {1990-06},
  journaltitle = {Journal of Child Language},
  shortjournal = {J Child Lang},
  volume = {17},
  number = {2},
  eprint = {2380274},
  eprinttype = {pubmed},
  pages = {357--374},
  issn = {0305-0009},
  doi = {10.1017/s0305000900013817},
  langid = {english}
}

@article{rastle358534NonwordsARC2002,
  title = {358,534 Nonwords: {{The ARC Nonword Database}}},
  shorttitle = {358,534 Nonwords},
  author = {Rastle, Kathleen and Harrington, Jonathan and Coltheart, Max},
  date = {2002},
  journaltitle = {The Quarterly Journal of Experimental Psychology Section A},
  volume = {55},
  number = {4},
  eprint = {12420998},
  eprinttype = {pubmed},
  pages = {1339--1362},
  issn = {0272-4987},
  doi = {10.1080/02724980244000099},
  url = {http://www.tandfonline.com/doi/abs/10.1080/02724980244000099},
  urldate = {2013-04-06}
}

@article{mezardConstraintSatisfactionProblems2009,
  title = {Constraint Satisfaction Problems and Neural Networks: {{A}} Statistical Physics Perspective},
  shorttitle = {Constraint Satisfaction Problems and Neural Networks},
  author = {Mézard, Marc and Mora, Thierry},
  date = {2009-01-01},
  journaltitle = {Journal of Physiology-Paris},
  shortjournal = {Journal of Physiology-Paris},
  series = {Neuromathematics of {{Vision}}},
  volume = {103},
  number = {1},
  pages = {107--113},
  issn = {0928-4257},
  doi = {10.1016/j.jphysparis.2009.05.013},
  url = {https://www.sciencedirect.com/science/article/pii/S0928425709000278},
  urldate = {2026-01-26}
}

@article{lupyan_meaningless_2015,
  title = {Meaningless Words Promote Meaningful Categorization},
  author = {Lupyan, G. and Casasanto, D.},
  date = {2015},
  journaltitle = {Language and Cognition},
  volume = {7},
  number = {2},
  pages = {167--193},
  doi = {10.1017/langcog.2014.21}
}

@book{landauLanguageExperienceEvidence1985,
  title = {Language and {{Experience}}: {{Evidence}} from the {{Blind Child}}},
  shorttitle = {Language and {{Experience}}},
  author = {Landau, Barbara and Gleitman, Lila R.},
  date = {1985},
  publisher = {Harvard University Press},
  location = {Cambridge, Mass},
  isbn = {978-0-674-51025-8},
  langid = {english},
  pagetotal = {250}
}

@incollection{diesselChapter3Usagebased2019,
  title = {Usage-Based Construction Grammar},
  shorttitle = {Chapter 3},
  booktitle = {Cognitive {{Linguistics}} - {{A Survey}} of {{Linguistic Subfields}}},
  author = {Diessel, Holger},
  editor = {Dąbrowska, Ewa and Divjak, Dagmar},
  date = {2019-07-08},
  pages = {50--80},
  publisher = {De Gruyter Mouton},
  url = {https://www.degruyterbrill.com/document/doi/10.1515/9783110626452-003/html?lang=en&srsltid=AfmBOor6xBHpZ35fC0NJkbSvcSdh-af7sCcWKhN-jKx9YawCS_GFG9Ks},
  urldate = {2026-02-03},
  isbn = {978-3-11-062645-2},
  langid = {english}
}

@book{croftRadicalConstructionGrammar2001,
  title = {Radical {{Construction Grammar}}: {{Syntactic Theory}} in {{Typological Perspective}}},
  shorttitle = {Radical {{Construction Grammar}}},
  author = {Croft, William},
  date = {2001},
  publisher = {Oxford University Press},
  isbn = {978-0-19-829954-7},
  langid = {english},
  pagetotal = {448}
}

@incollection{naiglesSyntacticSupportsWord2008,
  title = {Syntactic {{Supports}} for {{Word Learning}}},
  booktitle = {Blackwell {{Handbook}} of {{Language Development}}},
  author = {Naigles, Letitia R. and Swensen, Lauren D.},
  date = {2008},
  pages = {212--231},
  publisher = {John Wiley \& Sons, Ltd},
  doi = {10.1002/9780470757833.ch11},
  url = {http://onlinelibrary.wiley.com/doi/abs/10.1002/9780470757833.ch11},
  urldate = {2019-01-08},
  isbn = {978-0-470-75783-3},
  langid = {english}
}

@article{kakoSemanticsSyntacticStructures2001,
  title = {The Semantics of Syntactic Structures},
  author = {Kako, Edward and Wagner, Laura},
  date = {2001-03-01},
  journaltitle = {Trends in Cognitive Sciences},
  shortjournal = {Trends in Cognitive Sciences},
  volume = {5},
  number = {3},
  pages = {102--108},
  issn = {1364-6613},
  doi = {10.1016/S1364-6613(00)01594-1},
  url = {https://www.sciencedirect.com/science/article/pii/S1364661300015941},
  urldate = {2026-01-25}
}

@article{brown_linguistic_1957,
  title = {Linguistic Determinism and the Part of Speech},
  author = {Brown, R.},
  date = {1957},
  journaltitle = {Journal of Abnormal and Social Psychology},
  volume = {55},
  pages = {1--5}
}

@article{fisherSemanticContentSubcategorization1991,
  title = {On the Semantic Content of Subcategorization Frames},
  author = {Fisher, C. and Gleitman, H. and Gleitman, L. R.},
  date = {1991-07},
  journaltitle = {Cognitive Psychology},
  shortjournal = {Cogn Psychol},
  volume = {23},
  number = {3},
  eprint = {1884596},
  eprinttype = {pubmed},
  pages = {331--392},
  issn = {0010-0285},
  doi = {10.1016/0010-0285(91)90013-e},
  langid = {english}
}

@article{delucaRapidLearningWord2019,
  title = {Rapid Learning of Word Meanings from Distributional and Morpho-Syntactic Cues},
  author = {De Luca, Margherita and Lupyan, Gary},
  date = {2019},
  journaltitle = {Proceedings of the Annual Meeting of the Cognitive Science Society},
  volume = {41},
  number = {0},
  url = {https://escholarship.org/uc/item/5s50w37d},
  urldate = {2026-01-22},
  langid = {english}
}

@article{fisherDevelopmentalOriginsSyntactic2020,
  title = {The Developmental Origins of Syntactic Bootstrapping},
  author = {Fisher, Cynthia and Jin, Kyong-sun and Scott, Rose M.},
  date = {2020-01},
  journaltitle = {Topics in cognitive science},
  shortjournal = {Top Cogn Sci},
  volume = {12},
  number = {1},
  eprint = {31419084},
  eprinttype = {pubmed},
  pages = {48--77},
  issn = {1756-8757},
  doi = {10.1111/tops.12447},
  url = {https://pmc.ncbi.nlm.nih.gov/articles/PMC7004857/},
  urldate = {2026-01-16},
  pmcid = {PMC7004857}
}

@online{lupyanUnreasonableEffectivenessPattern2026,
  title = {The Unreasonable Effectiveness of Pattern Matching},
  author = {Lupyan, Gary and family=Arcas, given=Blaise Agüera, prefix=y, useprefix=false},
  date = {2026-01-16},
  eprint = {2601.11432},
  eprinttype = {arXiv},
  eprintclass = {cs},
  doi = {10.48550/arXiv.2601.11432},
  url = {http://arxiv.org/abs/2601.11432},
  urldate = {2026-01-21},
  pubstate = {prepublished}
}

@book{biberDimensionsRegisterVariation1995,
  title = {Dimensions of {{Register Variation}}: {{A Cross-Linguistic Comparison}}},
  shorttitle = {Dimensions of {{Register Variation}}},
  author = {Biber, Douglas},
  date = {1995},
  publisher = {Cambridge University Press},
  location = {Cambridge},
  doi = {10.1017/CBO9780511519871},
  url = {https://www.cambridge.org/core/books/dimensions-of-register-variation/FF817F2C32378B398C8019090381352E},
  urldate = {2026-02-01},
  isbn = {978-0-521-47331-6}
}

@unpublished{bojanowski_enriching_2016,
  title = {Enriching {{Word Vectors}} with {{Subword Information}}},
  author = {Bojanowski, Piotr and Grave, Edouard and Joulin, Armand and Mikolov, Tomas},
  date = {2016-07-15},
  eprint = {1607.04606},
  eprinttype = {arXiv},
  eprintclass = {cs},
  url = {http://arxiv.org/abs/1607.04606},
  urldate = {2017-04-05}
}

@article{reinhartLLMsWriteHumans2025,
  title = {Do {{LLMs}} Write like Humans? {{Variation}} in Grammatical and Rhetorical Styles},
  shorttitle = {Do {{LLMs}} Write like Humans?},
  author = {Reinhart, Alex and Markey, Ben and Laudenbach, Michael and Pantusen, Kachatad and Yurko, Ronald and Weinberg, Gordon and Brown, David West},
  date = {2025-02-25},
  journaltitle = {Proceedings of the National Academy of Sciences},
  volume = {122},
  number = {8},
  pages = {e2422455122},
  publisher = {Proceedings of the National Academy of Sciences},
  doi = {10.1073/pnas.2422455122},
  url = {https://www.pnas.org/doi/10.1073/pnas.2422455122},
  urldate = {2025-07-23}
}

\end{document}